\documentclass[11pt]{article} 
\usepackage{fullpage}
\usepackage{amsmath, amsfonts,amssymb}
\usepackage{graphicx,color,colortbl} 
\usepackage{subfigure}
\usepackage{tikz}
\usepackage{xcolor}
\usepackage{natbib}
\bibpunct{(}{)}{;}{a}{,}{,}

\usepackage{algorithm}
\usepackage{algorithmic}

\newcommand{\bbI}{\ensuremath{\mathbb{I}}}

\newcommand{\bbR}{\ensuremath{\mathbb{R}}}

\newtheorem{theorem}{Theorem}[section]

\DeclareMathOperator*{\KL}{KL}

\definecolor{LightGreen}{rgb}{0.56,0.93,0.56}
\definecolor{LightRed}{rgb}{0.93,0.56,0.56}
\definecolor{LightYellow}{rgb}{0.75,0.75,0.56}

\begin{document}

\title{Regularized Minimax Conditional Entropy for Crowdsourcing}

\author{
	Dengyong Zhou \thanks{Microsoft Research, Redmond, WA 98052. Email: {\tt dengyong.zhou@microsoft.com}.}
\and
    Qiang Liu \thanks{Department of Computer Science, University of California at Irvine, Irvine, CA 92637. Email: {\tt qliu1@uci.edu}.}
\and
John C. Platt  \thanks{Microsoft Research, Redmond, WA 98052. Email: {\tt jplatt@microsoft.com}.}
\and 
Christopher Meek \thanks{Microsoft Research, Redmond, WA 98052. Email: {\tt meek@microsoft.com}.}
\and
    Nihar B. Shah\thanks{Department of Electrical Engineering and Computer Science, University of California at Berkeley, Berkeley, CA 94720.  Email: {\tt nihar@eecs.berkeley.edu}. }
}

\date{}



\maketitle

\begin{abstract}
There is a rapidly increasing interest in  crowdsourcing for data labeling. By crowdsourcing,  a large number of labels can be often quickly gathered at low cost. However, the labels provided by the crowdsourcing workers are usually not of high quality. In this paper, we propose a minimax conditional entropy principle to infer ground truth from noisy crowdsourced labels. Under this principle, we derive a unique probabilistic labeling model jointly parameterized by worker ability and item difficulty.  We also propose an objective measurement principle,  and show that our  method is the only method which satisfies this objective measurement principle. We validate our method through  a variety of real crowdsourcing datasets with binary, multiclass or ordinal labels.
\end{abstract}

\newpage

\section{Introduction}
In many real-world applications,  the quality of a machine learning system is
governed by the number of labeled training examples, but the labor for data labeling is usually costly. There has been
considerable machine learning research work  on learning
when there are only few labeled examples,  such as semi-supervised learning and active learning.
In recent years,
with the emergence of crowdsourcing (or human computation) services like
Amazon Mechanical Turk\footnote{{https://www.mturk.com}}, the costs associated with collecting labeled data in many domains
have dropped dramatically enabling the collection of large amounts of
labeled data at a low cost. However, the labels provided by the
workers are often not of high quality, in part, due
to misaligned incentives and a lack of domain expertise in the workers.
To overcome this quality issue, in general, the items are redundantly labeled by several
different workers,  and then the workers' labels are aggregated in some manner, for example,  majority voting.

The assumption underlying  majority
voting is that all workers are equally good so they have equal vote. Obviously, such an assumption does not reflect the truth.  It is easy to imagine that one worker is more capable than another in some labeling task. More subtly, the skill level of a worker may significantly vary from one labeling category to another. To address these issues,
\citet{DawSke79} propose a model which assumes that each worker has a latent probabilistic confusion matrix for generating her
labels. The off-diagonal elements of the matrix represent the probabilities that the worker mislabels
an item from one class as another while the diagonal elements correspond to her accuracy in each
class. The true labels of the items and the confusion matrices of the workers
can be jointly estimated by maximizing the likelihood of the workers' labels.

 In the Dawid-Skene method, the performance of a worker characterized by her confusion matrix stays the same across  all items in the same class. That is not true in many labeling tasks, where some items are more difficult to label than others, and a worker is more likely to mislabel a difficult item than an easy one. Moreover, an item may be easily mislabeled as some class rather than others by whoever labels it. To address these issues, we develop a minimax conditional entropy principle for crowdsourcing. Under this principle, we derive a unique probabilistic model which takes both worker ability and item difficulty into account.  When item difficult is ignored, our model seamlessly reduces to the classical Dawid-Skene model. We also propose a natural objective measurement principle,  and show that our method is the only method which satisfies this objective measurement principle.

The work is an extension of the earlier results presented in \citep{zhoplaby12,ZhoLiuPlaMee14}. We organize the paper as follows. In Section \ref{sec:cat}, we propose the minimax conditional entropy principle for aggregating multiclass labels collected from a crowd and derive its dual form. In Section \ref{sec:reg}, we develop regularized minimax conditional entropy for preventing overfitting and generating probabilistic labels. In Section \ref{sec:objective}, we propose the objective measurement principle which also leads to the  probabilistic model derived from the minimax conditional entropy principle.  In Section \ref{sec:ordinal}, we extend our minimax conditional entropy method to ordinal labels, where we need to introduce a new assumption called adjacency confusability. In Section \ref{sec:implementation}, we present a simple yet efficient coordinate ascent method to solve the minimax program through its dual form and also a method for model selection. Related work are discussed in Section \ref{sec:related}. Empirical results on real crowdsourcing data with binary, multiclass or ordinal labels are reported in Section \ref{sec:exp}, and conclusion are presented in Section \ref{sec:conclusion}.

%
%
%

\section{Minimax Conditional Entropy Principle}
\label{sec:cat}

In this section, we present the minimax conditional entropy principle for aggregating crowdsourced multiclass labels in both its primal and dual forms. We also show that minimax conditional entropy is equivalent to minimizing  Kullback-Leibler (KL) divergence.

\subsection{Notation and Problem Setting}
Assume that there are a group of workers indexed by $i,$  a set of items indexed by $j,$ and a number of classes  indexed by $k$ or $c.$  Let $x_{ij}$ be the observed label that worker $i$ assigns to item $j, $  and $X_{ij}$  be the corresponding random variable. Denote by  $Q(Y_j = c)$  the unobserved true probability that item $j$ belongs to class $c.$ A special case is that $Q(Y_j = c) = 1$  and $Q(Y_j = k) = 0$ for any other class $k \neq c.  $ That is, the labels are deterministic. Denote by $P(X_{ij} = k |Y_j = c)$ the probability that worker $i$ labels item $j$ as class $k$ while the true label is $c. $  Our goal is to estimate the unobserved true labels from the noisy workers' labels.

\subsection{Primal Form}
Our approach is built upon two  four-dimensional tensors with the four dimensions corresponding to workers $i, $ items $j, $  observed labels $k, $ and  true labels $c.$  The first  tensor is referred to as the empirical confusion tensor of which  each element is given by   \[\widehat{\phi}_{ij}(c, k) = Q(Y_j = c) \bbI(x_{ij} = k) \]  to represent  an observed confusion from class $c$  to class $k$ by worker $i $ on item $j.$ The other tensor is referred to as the expected confusion tensor of which  each element is given by \[\phi_{ij}(c, k) = Q(Y_j = c) P(X_{ij} = k |Y_j = c) \] to represent an expected confusion from class $c$  to class $k$ by worker $i $ on item $j.$

\begin{figure}
\begin{minipage}[b]{0.50\linewidth}
\begin{center}
\newcolumntype{g}{>{\columncolor{LightGreen}}c}
\begin{tabular}{|c|c|g|c|c|c|c|}
\hline
 & item 1 & item 2 & $\cdots$ & item $n$    \\
\hline
worker 1 & $x_{11}$ & $x_{12}$ &$\cdots$  & $x_{1n}$\\
\rowcolor{LightRed}
\hline
worker 2 & $x_{21}$ & \cellcolor{LightYellow} $x_{22}$ &$\cdots$ & $x_{2n}$\\
\hline
$\cdots$ & $\cdots$ & $\cdots$ & $\cdots$  & $\cdots$ \\
\hline
worker $m$ & $x_{m1}$ & $x_{m2}$ &$\cdots$  & $x_{mn}$\\
\hline
\end{tabular}
\end{center}
\end{minipage}
\hfill
\begin{minipage}[b]{0.50\linewidth}
\begin{center}
\newcolumntype{g}{>{\columncolor{LightGreen}}c}
\begin{tabular}{|c|c|g|c|c|c|c|}
\hline
 & item 1 & item 2 & $\cdots$ & item $n$    \\
\hline
worker 1 & $\pi_{11}$ & $\pi_{12}$ &$\cdots$  & $\pi_{1n}$\\
\rowcolor{LightRed}
\hline
worker 2 & $\pi_{21}$ & \cellcolor{LightYellow}$\pi_{22}$ &$\cdots$ & $\pi_{2n}$\\
\hline
$\cdots$ & $\cdots$ & $\cdots$ & $\cdots$  & $\cdots$ \\
\hline
worker $m$ & $\pi_{m1}$ & $\pi_{m2}$ &$\cdots$  & $\pi_{mn}$\\
\hline
\end{tabular}
\end{center}
\end{minipage}
\caption{Left table: observed labels $x_{ij}$ provided by worker $i$ for item $j.$ Right table: underlying distributions $\pi_{ij}$ of  worker $i$  for generating a label for item $j.$  In our  approach, the rows and columns of of the unobserved right table  are constrained to match the rows and columns of the observed left table.}
\label{fig:cfm}
\end{figure}

We assume that the  labels of the items are independent. Thus, the entropy of the observed workers' labels conditioned on the true labels can be written as
\begin{align}
H(X|Y) = & -\sum_{j, c} Q(Y_j = c) \sum_{i,k} P(X_{ij} = k|Y_j = c) \log P(X_{ij} = k|Y_j = c). \nonumber
\end{align}
Both the distributions $P$ and $Q$ are unknown here. To attack this problem, we first consider a simpler problem: estimate $P$ when $Q$ is given. Then, we proceed to jointly estimating $P$  and $Q$ when both are unknown.

Given the true label distribution $Q$, we propose to estimate $P$ which generates the workers' labels  by
\begin{equation}
\label{eq:me}
 \max_{P} \quad    H(X|Y),
\end{equation}
subject to the worker and item  constraints (Figure \ref{fig:cfm})
\begin{subequations}
\label{eq:mct}
\begin{align}
& \sum_j \left[\phi_{ij}(c, k) - \widehat{\phi}_{ij}(c, k)\right] = 0, \ \forall i, k, c, \label{eq:mct1}\\
& \sum_i \left[\phi_{ij}(c, k) - \widehat{\phi}_{ij}(c, k)\right] = 0, \ \forall j, k, c, \label{eq:mct2}
\end{align}
\end{subequations}
plus the probability  constraints
\begin{subequations}
\label{eq:prob}
\begin{align}
& \sum_k P(X_{ij} = k|Y_j = c) = 1, \  \forall i, j, c, \label{eq:prob1}\\
&  \sum_c Q(Y_j = c) = 1,  \ \forall j, \label{eq:prob2}\\
&  Q(Y_j = c) \ge 0, \ \forall j, c. \label{eq:prob3}
 \end{align}
 \end{subequations}
The constraints in Equation \eqref{eq:mct1} enforce the expected confusion counts in the worker dimension to match their empirical counterparts. Symmetrically, the constraints in Equation \eqref{eq:mct2} enforce the expected confusion counts in the item dimension to match their empirical counterparts. An illustration of  empirical confusion tensors  is shown in Figure \ref{fig:empten}.

\begin{figure}
\begin{minipage}[b]{\linewidth}
\begin{center}
\begin{tabular}{|c|c|c|c|c|c|c|}
\hline
 & item 1 & item 2 & item 3 & item 4 & item 5 & item 6  \\
\hline
worker $1$ & $1$ & $2$ &$2$  & $1$ & $3$ &  $2$\\
\hline
worker $2$ & $2$ & $1$ & $2$ & $2 $ & $1 $ & $3 $\\
\hline
worker $3$ & $1$ & $1$ & $1$  & $2$  & $2$ & $3$\\
\hline
\end{tabular}
\end{center}
\vskip 0.10in
\begin{equation*}
\widehat{\phi}_1 =
  \begin{pmatrix}
   1 & 1 & 0 \\
   1 & 1 & 0 \\
   0 & 1 & 1
  \end{pmatrix}, \quad
\widehat{\phi}_2 =
  \begin{pmatrix}
   1 & 1 & 0 \\
   0 & 2 & 0 \\
   1 & 0 & 1
  \end{pmatrix}, \quad
  \widehat{\phi}_3 =
  \begin{pmatrix}
   2 & 0 & 0 \\
   1 & 1 & 0 \\
   0 & 1 & 1
  \end{pmatrix}
\end{equation*}
\end{minipage}
\caption{An illustration of the empirical confusion tensors. The table contains three workers' labels over six items. These items are assumed to have deterministic true labels as follows: class 1 = \{item 1, item 2\}, class 2 = \{item 3, item 4\}, and class 3 = \{item 5, item 6\}. The $(c, k)$-th entry of matrix $\widehat{\phi}_i$ represents the number of the items labeled as class $k$ by worker $i$ given that  their true labels are class $c.$}
\label{fig:empten}
\end{figure}

When both the distributions $P$ and $Q$ are unknown, we propose to jointly estimate them by
\begin{equation}
\label{eq:mme}
\min_{Q} \max_{P} \quad    H(X|Y),
\end{equation}
subject to the constrains in Equation \eqref{eq:mct} and \eqref{eq:prob}.  Intuitively, entropy can be understood as a measure of uncertainty. Thus, minimizing the maximum conditional entropy means that, given the true labels,  the workers's labels are the least random. Theoretically,  minimizing the maximum conditional entropy  can be connected  to maximum likelihood. In what follows, we show how the connection is established.

\subsection{Dual Form}
\label{sec:mle}
The Lagrangian  of the maximization problem in \eqref{eq:mme} can be written as
\begin{equation}
\label{eq:lag1}
L = H(X|Y) + L _{\sigma} + L_{\tau}+ L_\lambda
\end{equation}
with
\begin{align}
L_{\sigma} = & \sum_{i, c, k} \sigma_i(c, k) \sum_j \left[\phi_{ij}(c, k) -  \widehat{\phi}_{ij}(c, k)\right] \nonumber,\\
L_{\tau} = &  \sum_{j, c, k} \tau_j(c, k)  \sum_i \left[\phi_{ij}(c, k) -  \widehat{\phi}_{ij}(c, k)\right] \nonumber, \\
L_{\lambda} = & \sum_{i,j, c} \lambda_{ijc}\bigg[\sum_{k} P(X_{ij} = k|Y_j = c) -1 \bigg],  \nonumber
\end{align}
where $\sigma_i(c, k), \tau_j(c, k)$ and $\lambda_{ijc}$ are introduced as the Lagrange multipliers.  By the Karush-Kuhn-Tucker (KKT) conditions \citep{bova04},
\begin{equation*}
 \frac{\partial L }{\partial P(X_{ij} = k|Y_j = c)} = 0,
\end{equation*}
which implies
\begin{equation*}
\log P(X_{ij} = k|Y_j = c) = \lambda_{ijc} -1 + \sigma_i(c, k)  + \tau_{j}(c, k).
\end{equation*}
Combining the above equation  and  the probability constraints in \eqref{eq:prob1}  eliminates $\lambda$ and yields
\begin{equation}
\label{eq:model}
P(X_{ij} = k|Y_j = c) = \frac{1}{Z_{ij}} \exp[\sigma_i(c, k) + \tau_{j}(c, k)],
\end{equation}
where $Z_{ij}$ is the normalization factor given by
\[Z_{ij} = \sum_k \exp[\sigma_i(c, k) + \tau_{j}(c, k)]. \]
Although the matrices $[\sigma_i(c, k)]$ and $[\tau_{j}(c, k)]$ in Equation \eqref{eq:model} come out as the mathematical consequence of minimax conditional entropy, they can be understood intuitively. We can consider the matrix $[\sigma_i(c, k)]$ as the measure of the intrinsic ability of worker $i.$ The $(c, k)$-th entry measures  how likely worker $i$ labels a randomly chosen item in class $c$ as class $k.$ Similarly, we can consider  the matrix $[\tau_i(c, k)]$ as the measure of the intrinsic  difficult of item $j.$ The $(c, k)$-th entry measures  how likely item $j$ in class $c$  is labeled as class $k$ by a randomly chosen worker. In the following, we refer to $[\sigma_i(c, k)]$ as  worker confusion matrices and $[\tau_i(c, k)]$ as item confusion matrices.

Substituting the labeling model in Equation \eqref{eq:model} into the Lagrangian in Equation \eqref{eq:lag1}, we can obtain the dual form of the minimax problem \eqref{eq:mme} as (see Appendix \ref{app:dual1})
\begin{equation}
\label{eq:dual1}
\max_{\sigma, \tau, Q} \quad  \sum_{j,c}Q(Y_j = c) \sum_i \log P(X_{ij} = x_{ij}|Y_j = c).
\end{equation}
 It is obvious  that, to be optimal, the true label distribution  has to be deterministic. Thus, the dual Lagrangian can be equivalently expressed as the complete log-likelihood
\begin{align}
& \log \bigg\{\prod_j \sum_{c} Q(Y_j = c) \prod_i P(X_{ij} = x_{ij}|Y_j = c)\bigg\}.  \nonumber
\end{align}
In Section \ref{sec:reg},  we show how to regularize  the objective function in \eqref{eq:mme} to generate probabilistic labels.

\subsection{Minimizing KL Divergence}
\label{sec:kl}
Let us extend the two distributions $P$ and $Q$  to the product space $X \times Y.$  We extend the distribution $Q$ by defining $ Q(X_{ij} = x_{ij}) = 1,  $  and  $Q(Y)$ stays the same. We extend  the  distribution $P$  with  $P(X, Y) = \prod_{ij}P{(X_{ij}|Y_j)} P(Y_j), $ where $P{(X_{ij}|Y_j)}$ is given by Equation \eqref{eq:model},   and  $P(Y)$ is a uniform distribution over all possible classes. Then,  we have
\begin{theorem}
\label{eq:minkl}
When the true labels are deterministic, minimizing the KL divergence from $Q$ to $P, $ that is,
\begin{equation}
\begin{aligned}
\min_{P, Q}  \bigg\{D_{\KL} (Q\parallel P) = \sum_{X, Y}Q(X,Y) \log \frac{Q(X,Y)}{P(X,Y)}\bigg\},
\end{aligned}
\end{equation}
is equivalent to the minimax problem in \eqref{eq:mme}.
\end{theorem}
The proof is presented in Appendix \ref{sec:proofkl}. A sketch of the proof is as follows. We show that,
\begin{equation*}
\begin{aligned}
D_{\KL}(Q\parallel P) = & - \sum_{j, c} Q(Y_j = c) \sum_{i, k} P(X_{ij} = k|Y_j = c)\log P(X_{ij} = k|Y_j = c) \\
 &  + \sum_{Y} Q(Y)\log Q(Y) -  \log P(Y). \\
\end{aligned}
\end{equation*}
By the definition of $P(X, Y), $ $P(Y)$ is a constant. Moreover, when the true labels are deterministic, we have
\[\sum_{Y} Q(Y)\log Q(Y) = 0. \]
This concludes the proof of this theorem.


\section{Regularized Minimax Conditional Entropy}
\label{sec:reg}

In this section, we regularize our minimax conditional entropy method to address two practical issues:
\begin{itemize}
\item \textbf{Preventing overfitting}. While crowdsourcing is cheap, collecting  many redundant labels may be more expensive than hiring experts. Typically, the number of labels collected for each item is  limited to a small number. In this case,  the empirical counts in Equation \eqref{eq:mct} may not match their expected values. It is likely that they fluctuate around their expected values although these fluctuations are not large.

\item \textbf{Generating probabilistic labels}. Our minimax conditional entropy method can only generate deterministic labels (see Section \ref{sec:mle}).  In practice,   probabilistic labels are usually  more useful than  deterministic  labels. When the estimated label distribution for an item is close to  uniform over several classes, we can either ask for more labels for the item from the crowd or forward the item to an external expert.
\end{itemize}

For addressing the issue of overfitting, we formulate our observation by replacing exact matching with approximate matching while penalizing large fluctuations. For generating probabilistic labels,  we consider an entropy regularization over the unknown true label distribution. This is motivated by the analysis in Section \ref{sec:kl}.

Formally, we regularize our minimax conditional entropy method as follows.  Let us denote the entropy of the true label distribution by
 \[ H(Y) = -\sum_{j, c} Q(Y_j = c)\log  Q(Y_j = c). \]
To estimate the true labels, we consider
\begin{equation}
\label{eq:rmme}
\min_{Q} \max_{P} \ H(X|Y) - H(Y) - \frac{1}{\alpha}\Omega(\xi)  - \frac{1}{\beta}\Psi(\zeta)
\end{equation}
subject to the relaxed  worker and item constraints
\begin{subequations}
\label{eq:xizeta}
\begin{align}
& \sum_j \left[\phi_{ij}(c, k) - \widehat{\phi}_{ij}(c, k)\right] = \xi_{i}(c, k),  \  \forall i,s, \label{eq:rmctw}\\
& \sum_i \left[\phi_{ij}(c, k) - \widehat{\phi}_{ij}(c, k)\right] = \zeta_{j}(c, k),  \ \forall j,s, \label{eq:rmcti}
\end{align}
\end{subequations}
plus the probability constraints in Equation \eqref{eq:prob}. The regularization functions $\Omega$ and $\Psi$ are chosen as
\begin{subequations}
\label{eq:ref}
 \begin{align}
& \Omega(\xi) =  \frac{1}{2}\sum_{i}\sum_{c, k} \left[\xi_{i}(c, k)\right]^2, \label{eq:refunome}\\
& \Psi(\zeta) =  \frac{1}{2} \sum_{j}\sum_{c, k} \left[\zeta_{j}(c, k)\right]^2. \label{eq:refunpsi}
\end{align}
\end{subequations}
The new slack variables $\xi_{i}(c, k), \zeta_{j}(c, k)$ in Equation \eqref{eq:xizeta}  model the possible fluctuations. Note that these slack variables  are not restricted to be positive. When there are a sufficiently large number of observations, the fluctuations should be approximately normally distributed, due to the central limit theorem. This observation motivates the choice of the regularization functions in \eqref{eq:ref} to penalize large fluctuations. The entropy term $H(Y)$ in the objective function,  which is introduced for generating probabilistic labels,  can be regarded as penalizing a large deviation from the uniform distribution.

 Substituting the labeling model from Equation \eqref{eq:model} into the Lagrangian of \eqref{eq:rmme}, we obtain the dual form (see Appendix \ref{app:dual2})
 \begin{align}
 \max_{\sigma, \tau, Q} \quad & \sum_{j,c}Q(Y_j = c) \sum_i \log P(X_{ij} = x_{ij}|Y_j = c) + H(Y) - {\alpha} \Omega^*(\sigma) - {\beta}\Psi^*(\tau),   \label{eq:dual2}
 \end{align}
 where
\begin{align}
& \Omega^*(\sigma) =  \frac{1}{2} \sum_{i}\sum_{c, k} \left[\sigma_{i}(c, k)\right]^2, \label{eq:cjomg}\\
& \Psi^*(\tau) = \frac{1}{2} \sum_{j}\sum_{c, k} \left[\tau_{j}(c, k)\right]^2. \label{eq:cjpsi}
\end{align}
 When $\alpha = 0$ and $ \beta = 0, $ the objective function in \eqref{eq:dual2} turns out to be a lower bound of the log marginal likelihood
 \begin{align}
& \log \bigg\{\prod_j \sum_{c} \prod_i P(X_{ij} = x_{ij}| Y_j = c)\bigg\}  \nonumber \\
= &   \log \bigg\{\prod_j \sum_{c} \frac{Q(Y_j = c)}{Q(Y_j = c)} \prod_i P(X_{ij} = x_{ij}| Y_j = c)\bigg\} \nonumber\\
\geq &  \sum_{j,c}Q(Y_j = c) \sum_i \log P(X_{ij} = x_{ij}| Y_j = c) + H(Y). \nonumber
\end{align}
The last step is based on Jensen's inequality. Maximizing the marginal likelihood is  more appropriate than maximizing the complete likelihood since only the observed data matters in our inference.

Finally, we introduce a variant of our regularized minimax conditional entropy. It is obtained by restricting the feasible region of the slack variables through
\begin{align}
\sum_{c} \xi_i(c, c) = 0, \ \forall i.   \label{eq:sumtozero}
\end{align}
This is equivalent to
\[\sum_{j,c} \left[\phi_{ij}(c, c) - \widehat{\phi}_{ij}(c, c)\right] = 0,  \  \forall i. \]
It says that, the empirical count of the correct answers from each worker is equal to its expectation. According to the law of large numbers,  this assumption is approximately correct when a worker has a sufficiently large number of correct answers.  Note that this does not mean that the percentage of the correct answers  from the worker has to be large. Let $K$ denote the class size. Under the additional constraints in Equation \eqref{eq:sumtozero}, the dual problem can still be expressed by \eqref{eq:dual2} except (see Appendix \ref{app:dual2})
\begin{equation}
\Omega^*(\sigma) =   \frac{1}{2} \sum_{i, c} \bigg(\left[\sigma_i(c, c) - \overline{\sigma_i(c, c)}\right]^2 +  \sum_{k \neq c}\left[\sigma_i(c, k) - \overline{\sigma_i(c, k)}\right]^2 \bigg),  \label{eq:comg} \\
\end{equation}
where
\[\overline{\sigma_i(c, c)} = \frac{1}{K}\sum_c\sigma_i(c, c), \quad  \overline{\sigma_i(c, k)} = \frac{1}{K(K-1)} \sum_c \sum_{k \neq c} \sigma_i(c, k). \]
From our empirical evaluations, this variant is somewhat worse than its original version on most datasets. We include it here only for theoretic interest.

\section{Objective Measurement Principle}

\label{sec:objective}

In this section, we introduce a natural objective measurement principle, and show that the probabilistic labeling model in Equation \eqref{eq:model} is a consequence of this principle.

Intuitively, the objective measurement principle can be described as follows:
\begin{enumerate}
\item  A comparison of labeling difficulty between two items should be independent of which particular workers were involved in the comparison; and it should also be independent of which other items  might also be compared.
\item  Symmetrically, a comparison of labeling ability  between two workers should be independent of which particular items  were involved  in the comparison; and it should also be independent of which other workers might also be compared.
\end{enumerate}
Next we  mathematically define the objective measurement principle.

Assume that worker $i$ has labeled items $j$ and $j'$ in class $c.$  Denote by $E$ the event that one of these two items is labeled as $k, $ and the other is labeled as $c.$ Formally,
\[E = \left\{\bbI(X_{ij} = k) + \bbI(X_{ij'} = k) = 1, \ \bbI(X_{ij} = c) + \bbI(X_{ij'} = c) = 1\right\}.\]
Denote by $A$ the event that item $j$ is labeled as $k$ and item $j'$ is labeled as $c.$ Formally,
\[A = \left\{X_{ij} = k, \ X_{ij'} = c \right\}. \]
It is obvious that $A \subset E. $ Now we formulate the requirement (1) in the objective measurement principle as follows:  $P(A|E)$ is independent of worker $i$.  Note that
\begin{align*}
P(A|E) = \frac{P(X_{ij} = k|Y_j = c)P(X_{ij'} = c|Y_{j'} = c)}{P(X_{ij} = k|Y_j = c)P(X_{ij'} = c|Y_{j'} = c) + P(X_{ij} = c|Y_j = c)P(X_{ij'} = k|Y_{j'} = c)}.
\end{align*}
Hence, $P(A|E)$ is independent of worker $i$ if and only if
\[\frac{P(X_{ij} = k|Y_j = c)P(X_{ij'} = c|Y_{j'} = c)}{P(X_{ij} = c|Y_j = c)P(X_{ij'} = k|Y_{j'} = c)}\]
is independent of worker $i$. In other words, given another arbitrary worker $i',$ we should have
\[\frac{P(X_{ij} = k|Y_j = c)P(X_{ij'} = c|Y_{j'} = c)}{P(X_{ij} = c|Y_j = c)P(X_{ij'} = k|Y_{j'} = c)}  = \frac{P(X_{i'j} = k|Y_j = c)P(X_{i'j'} = c|Y_{j'} = c)}{P(X_{i'j} = c|Y_j = c)P(X_{i'j'} = k|Y_{j'} = c)}.   \]
Without loss of generality, we choose $i' = 0, \ j' = 0 $ as the fixed references. Then,
\[\frac{P(X_{ij} = k|Y_j = c)}{P(X_{ij} = c|Y_j = c)} \propto \frac{P(X_{i0} = k|Y_{0} = c)}{P(X_{i0} = c|Y_{0} = c)}\frac{P(X_{0j} = k|Y_j = c)}{P(X_{0j} = c|Y_j = c)}. \]
By the fact that probabilities are nonnegative, we can write
\begin{align*}
{P(X_{i0} = k|Y_{0} = c)} = \exp[\sigma_i(c, k)], \quad {P(X_{0j} = k|Y_j = c)} = \exp[\tau_j(c, k)].
 \end{align*}
The probabilistic labeling model in  Equation \eqref{eq:model} follows immediately. It is easy to verify that due to the symmetry between item difficulty and worker ability, we can instead start from formulating the requirement (2) in the objective measurement principle to achieve the same result. Hence, in this sense, the two requirements  are actually redundant.

\section{Extension to Ordinal Labels}
\label{sec:ordinal}

In this section, we extend the minimax conditional entropy principle from multiclass to ordinal labels.  Eliciting ordinal labels is important in tasks such as judging the relative quality of web search results or consumer products. Since ordinal labels are a special case of  multiclass labels, the approach that we have developed in the previous sections can be used to aggregate ordinal labels. However, we observe that,  in ordinal labeling,  workers usually have an error pattern  different from what we observe in multiclass labeling. We summarize our observation as  the adjacency confusability assumption, and  formulate it by introducing a different set of  constraints for workers and items.

\subsection{Adjacency Confusability}
In ordinal labeling, workers usually have difficulty distinguishing between two adjacent ordinal classes whereas distinguishing between two classes which are far away from each other is much easier. We refer to this observation as adjacency confusability.

To illustrate this observation, let us consider the example of screening mammograms. A mammogram is an x-ray picture used to check for breast cancer in women. Radiologists often rate mammograms on a scale such as no cancer, benign cancer, possible malignancy, or malignancy.  In screening mammograms, a radiologist may rate a mammogram which indicates possible malignancy as malignancy, but it is less likely  that she rates a mammogram which  indicates  no cancer as  malignancy.

\subsection{Ordinal Minimax Conditional Entropy}
 In what follows, we construct a different set of worker and item constraints  to encode adjacency confusability. The formulation leads to an ordinal labeling  model parameterized with  \emph{structured} confusion matrices for workers and items.

We introduce two symbols  $\Delta$ and $\nabla$ which take on arbitrary binary relations in $\{\ge, <\}. $  Ordinal labels are represented by consecutive integers, and the minima one is $0.$ To estimate the true ordinal labels, we consider
\begin{equation}
\label{eq:omme}
\min_{Q} \max_{P} \ H(X|Y)
\end{equation}
subject to the ordinal-based worker and item constraints
\begin{subequations}
\label{eq:omct}
\begin{align}
& \sum_{ c  \Delta s} \sum_{k \nabla s } \sum_j \left[\phi_{ij}(c, k) - \widehat{\phi}_{ij}(c, k)\right] = 0, \  \forall i,s \ge 1, \label{eq:omctw}\\
& \sum_{ c  \Delta s} \sum_{k \nabla s } \sum_i \left[\phi_{ij}(c, k) - \widehat{\phi}_{ij}(c, k)\right] = 0,  \ \forall j,s \ge 1, \label{eq:omcti}
\end{align}
\end{subequations}
for all $\Delta, \nabla \in \{\ge, <\}, $  and the probability constraints in \eqref{eq:prob}. We exclude the case $s=0$ in which the constraints trivially  hold.

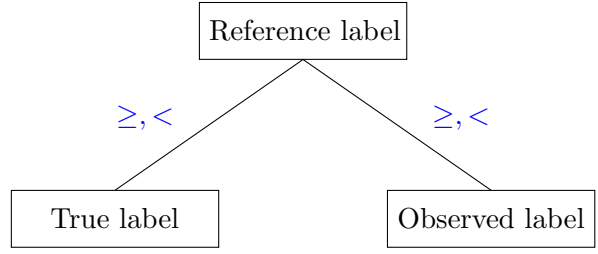
\begin{figure}[t]
\centering    
\begin{tikzpicture}
\tikzstyle{gbox}=[draw=black, shape=rectangle, text centered, fill=white];
\node[gbox,  text width=2.5cm, text height= 0.5cm]   at (5.5,3.5) {\raisebox{.12cm}{Reference label}}; 
\node[gbox,  text width=2.5cm, text height= 0.5cm]  at (3,1) {\raisebox{.12cm}{True label}};
\node[gbox,  text width=2.5cm, text height= 0.5cm]  at (8,1) {\raisebox{.12cm}{Observed label}};
\draw (5.5,3.12) -- (3, 1.38);
\draw (5.5,3.12) -- (8, 1.38);
\node  (a5)  at  (3.4,2.35) {\textcolor{blue}{$\geq, <$} };
\node  (a5)  at  (7.6,2.35) {\textcolor{blue}{$\geq, <$} };
\end{tikzpicture}
\caption{Indirect comparison between a true label and an observed label via comparing both to a {reference label} which varies through all possible values  in a given ordinal label set. }
\label{fig:ord}
\end{figure}

Let us explain the meaning of the constraints in Equation \eqref{eq:omct}. To construct ordinal-based constraints, the first issue that we have to address is how to compare the observed label $x_{ij}$ and the true label $Y_j$ in an ordinal sense. For multiclass labels, as we have seen in Section \ref{sec:cat}, the label comparison problem is trivial: we only need to check whether they are equal or not. For ordinal labels,  such a problem becomes tricky. Here,  we propose an indirect comparison between two ordinal labels by comparing both to a \emph{reference label} $s$ which varies through all possible values  in a given ordinal label set (Figure \ref{fig:ord}).  Consequently, for every chosen reference label $s, $ we partition the Cartesian  product of the label set into four disjoint regions
\begin{align}
& \{(c, k)| c < s, k < s\}, \  \{(c, k)| c < s, k \ge s\}, \nonumber\\
& \{(c, k)| c \ge s, k < s\}, \ \{(c, k)| c \ge s, k \ge s\}.  \nonumber
\end{align}
A partition example is shown in Table 1 where the given label set is $\{0,1,2,3\}.$  Then, Equation \eqref{eq:omctw} defines a set of constraints for the workers by summing Equation \eqref{eq:mct1} over each region. Similarly, Equation \eqref{eq:omcti} defines a set of constraints for the items  by summing Equation \eqref{eq:mct2} over each region.

From the discussion above, we can see that when there are more than two ordinal classes, the constraints in Equation \eqref{eq:omct} are less restrictive than those in Equation \eqref{eq:mct}. Consequently, as we see below, the labeling model resulted from Equation \eqref{eq:omct} has fewer parameters. In the case in which there are only two ordinal classes, the  sets of disjoint regions degenerate to pairs $(c,k)$ and, thus, the sets of constraints in Equations \eqref{eq:omct} and \eqref{eq:mct} are identical.


Next we explain why we construct the ordinal-based constraints in such a way. 
Let us write
\begin{align}
\sum_{ c  \Delta s} \sum_{k \nabla s } \sum_j \widehat{\phi}_{ij}(c, k) & = \sum_j \sum_{ c \Delta s}  \sum_{k \nabla s } Q(Y_j = c) \bbI(x_{ij} = k) \nonumber \\
& = \sum_j \sum_{ c \Delta s} Q(Y_j = c) \sum_{k \nabla s } \bbI(x_{ij} = k)\nonumber\\
& = \sum_j Q(Y_j \Delta s) \bbI(x_{ij} \nabla s). \nonumber
\end{align}
For example, when $\Delta = <$ and $\nabla = \ge, $ the above equation becomes
\[\sum_j \sum_{ c  < s} \sum_{k \ge s } \widehat{\phi}_{ij}(c, k) = \sum_j Q(Y_j < s) \bbI(x_{ij} \ge s). \]
This counts the items of which each belongs to a class less than $s$ but worker $i $  assigned a label larger or equal to $s.$

In general,
for a comparison between an observed label  and a reference label, there are two possible outcomes: the observed label is larger or equal to the reference label; or the observed label is smaller than the reference label. These are also the two possible outcomes for a comparison between a true label  and a reference label. Putting these together, we have four  possible outcomes in total.
The constraints in Equation \eqref{eq:omctw} enforce expected counts of all the four kinds of outcomes in the worker dimension to match their empirical counterparts. Symmetrically, the constraints in Equation \eqref{eq:omcti} enforce  expected counts of all the four kinds of outcomes in the item dimension to match their empirical counterparts.

\begin{table}
\centering
\subtable[Partitioning with $s=1$]{
 \begin{tabular}{c|ccc}
 {$(0, 0)$} & $(0, 1)$ &  $(0, 2)$ & $(0, 3)$       \\ \hline
 {$(1, 0)$} & $(1, 1)$ & $(1, 2)$ & $(1, 3)$     \\
 {$(2, 0)$} & $(2, 1)$ & $(2, 2)$ &  $(2, 3)$       \\
 {$(3, 0)$} & $(3, 1)$ &  $(3, 2)$ &  $(3, 3)$
 \end{tabular}
 }
 \vskip 0.13in
\subtable[Partitioning with $s=2$]{
 \begin{tabular}{cc|cc}
 {$(0, 0)$} & $(0, 1)$ &  $(0, 2)$ & $(0, 3)$       \\
 {$(1, 0)$} & $(1, 1)$ & $(1, 2)$ & $(1, 3)$     \\ \hline
 {$(2, 0)$} & $(2, 1)$ & $(2, 2)$ &  $(2, 3)$       \\
 {$(3, 0)$} & $(3, 1)$ &  $(3, 2)$ &  $(3, 3)$
 \end{tabular}
}
\vskip 0.13in
\subtable[Partitioning with $s=3$]{
 \begin{tabular}{ccc|c}
 {$(0, 0)$} & $(0, 1)$ &  $(0, 2)$ & $(0, 3)$       \\
 {$(1, 0)$} & $(1, 1)$ & $(1, 2)$ & $(1, 3)$     \\
 {$(2, 0)$} & $(2, 1)$ & $(2, 2)$ &  $(2, 3)$       \\ \hline
 {$(3, 0)$} & $(3, 1)$ &  $(3, 2)$ &  $(3, 3)$
 \end{tabular}
}
\caption{Partitioning the Cartesian product of the ordinal label set $\{0, 1, 2, 3\}.$ With respect to each possible reference label,  each table is partitioned into four disjoint regions.   }
\label{fig:sp}
\end{table}

The Lagrangian of the maximization  problem in \eqref{eq:omme}  can be written as
\begin{equation*}
 L  =  H(X|Y)  + L_{\sigma} + L_{\tau}  + L_{\lambda},
\end{equation*}
with
\begin{equation*}
\begin{aligned}
 L_{\sigma}  = & \sum_{i, s} \sum_{\Delta, \nabla}\sigma^{{\Delta, \nabla}}_{is} \sum_{ c  \Delta s} \sum_{k \nabla s } \sum_j \left[\phi_{ij}(c, k) - \widehat{\phi}_{ij}(c, k)\right],\\
 L_{\tau}    = & \sum_{j, s} \sum_{\Delta, \nabla} \tau^{\Delta, \nabla}_{js} \sum_{ c  \Delta s} \sum_{k \nabla s } \sum_i \left[\phi_{ij}(c, k) - \widehat{\phi}_{ij}(c, k)\right], \\
 L_{\lambda} = &  \sum_{i,j, c} \lambda_{ijc}\bigg[\sum_k P(X_{ij} = k|Y_j = c) -1\bigg],
\end{aligned}
\end{equation*}
where $\sigma^{{\Delta, \nabla}}_{is}, \tau^{\Delta, \nabla}_{js}$ and $\lambda_{ijc}$ are the introduced Lagrange multipliers. By a procedure similar to that in Section \ref{sec:cat}, we obtain a probabilistic ordinal labeling model
\begin{equation}
\label{eq:rating}
P(X_{ij} = k|Y_j = c) = \frac{1}{Z_{ij}} \exp[ \sigma_i (c, k) + \tau_j (c, k)], 
\end{equation}
where
\begin{subequations}
\label{eq:osigmatau}
\begin{align}
\sigma_{i}(c, k)  = & \sum_{s \geq 1} \sum_{\Delta, \nabla} \sigma^{\Delta, \nabla}_{is}\bbI(c \Delta s, k \nabla s), \label{eq:osigma} \\
\tau_{j}(c, k)  = & \sum_{s \geq 1} \sum_{\Delta, \nabla}  \tau^{\Delta, \nabla}_{js}\bbI(c \Delta s, k \nabla s). \label{eq:otau}
\end{align}
\end{subequations}
The ordinal labeling  model in Equation \eqref{eq:rating} is actually the same as  the  multiclass labeling model in Equation \eqref{eq:model} except  the worker and item confusion matrices in Equation \eqref{eq:rating} are  now subtly structured through Equation \eqref{eq:osigmatau}. It is because of the structure that the ordinal labeling  model  has  fewer parameters than the multiclass labeling model when there are more than two classes.  In the case in which there are only two classes, the ordinal labeling model and the multiclass labeling model coincide as one would expect.

The regularized minimax conditional entropy for ordinal labels can be written as
\begin{equation}
\label{eq:ormme}
\min_{Q} \max_{P} \ H(X|Y) - H(Y) - \frac{1}{\alpha}\Omega(\xi)  - \frac{1}{\beta}\Psi(\zeta)
\end{equation}
subject to the relaxed  worker and item constraints
\begin{subequations}
\label{eq:oxizeta}
\begin{align}
& \sum_{ c  \Delta s} \sum_{k \nabla s } \sum_j \left[\phi_{ij}(c, k) - \widehat{\phi}_{ij}(c, k)\right] = \xi_{is}^{\Delta, \nabla} ,   \forall i,s, \label{eq:ormctw}\\
& \sum_{ c  \Delta s} \sum_{k \nabla s } \sum_i \left[\phi_{ij}(c, k) - \widehat{\phi}_{ij}(c, k)\right] = \zeta_{js}^{\Delta, \nabla},  \forall j,s, \label{eq:ormcti}
\end{align}
\end{subequations}
for all $\Delta, \nabla \in \{\ge, <\}, $ and the probability constraints in Equation \eqref{eq:prob}. When we choose
 \begin{align*}
& \Omega(\xi) = \frac{1}{2}\sum_{i,s}\sum_{\Delta, \nabla} \left(\xi_{is}^{\Delta, \nabla}\right)^2, \\
& \Psi(\zeta)  = \frac{1}{2}\sum_{j,s}\sum_{\Delta, \nabla} \left(\zeta_{js}^{\Delta, \nabla}\right)^2,
\end{align*}
the dual problem becomes
\begin{align*}
 \max_{\sigma, \tau, Q} \quad & \sum_{j,c}Q(Y_j = c) \sum_i \log P(X_{ij} = x_{ij}|Y_j = c) +  H(Y) - \alpha \Omega^*(\sigma) - \beta \Psi^*(\tau),   \label{eq:odual}
 \end{align*}
 where
  \begin{align*}
& \Omega^*(\sigma) = \frac{1}{2}\sum_{i,s}\sum_{\Delta, \nabla} \left(\sigma_{is}^{\Delta, \nabla}\right)^2, \\
& \Psi^*(\tau)  = \frac{1}{2}\sum_{j,s}\sum_{\Delta, \nabla} \left(\tau_{js}^{\Delta, \nabla}\right)^2.
\end{align*}
\subsection{Ordinal Objective Measurement Principle}
In this section, we adapt the objective measurement principle developed in Section \ref{sec:objective} to ordinal labels.

Assume that worker $i$ has labeled items $j$ and $j'$ in class $c.$  For any class $k, $ we define two events. The first event is
\[E = \left\{\bbI(X_{ij} = k) + \bbI(X_{ij'} = k) = 1, \ \bbI(X_{ij} = k+1) + \bbI(X_{ij'}= k+1) = 1\right\}, \]
and the other event is
\[A = \left\{X_{ij} = k, \ X_{ij'}  = k+1 \right\}. \]
Note that $A \subset E. $ Now we formulate the objective measurement principle as follows:  $P(A|E)$ is independent of worker $i$. Assume that the labels of the items are independent. Then,  $P(A|E)$ can be written as
\begin{align*}
\frac{P(X_{ij} = k|Y_j = c)P(X_{ij'} = k+1|Y_{j'} = c)}{P(X_{ij} = k|Y_j = c)P(X_{ij'} = k+1|Y_{j'} = c) + P(X_{ij} = k+1|Y_j = c)P(X_{ij'} = k|Y_{j'} = c)}.
\end{align*}
Hence, $P(A|E)$ is independent of worker $i$ if and only if
\[\frac{P(X_{ij} = k|Y_j = c)P(X_{ij'} = k+1|Y_{j'} = c)}{P(X_{ij} = k+1|Y_j = c)P(X_{ij'} = k|Y_{j'} = c)}\]
is independent of worker $i$. In other words, given another arbitrary worker $i',$ we should have
\[\frac{P(X_{ij} = k|Y_j = c)P(X_{ij'} = k+1|Y_{j'} = c)}{P(X_{ij} = k+1|Y_j = c)P(X_{ij'} = k|Y_{j'} = c)}  = \frac{P(X_{i'j} = k|Y_j = c)P(X_{i'j'} = k+1|Y_{j'} = c)}{P(X_{i'j} = k+1|Y_j = c)P(X_{i'j'} = k|Y_{j'} = c)}.   \]
To introduce adjacency confusability, we further assume that, for any two classes $c, c' \ge k+1$ (or $c, c' < k+1$),
\[\frac{P(X_{ij} = k|Y_j = c)P(X_{ij'} = k+1|Y_{j'} = c)}{P(X_{ij} = k+1|Y_j = c)P(X_{ij'} = k|Y_{j'} = c)}  = \frac{P(X_{ij} = k|Y_j = c')P(X_{ij'} = k+1|Y_{j'} = c')}{P(X_{ij} = k+1|Y_j = c')P(X_{ij'} = k|Y_{j'} = c')}. \]
Then,  by a procedure similar to that in Section \ref{sec:objective}, we  reach the probabilistic ordinal labeling model described by Equation \eqref{eq:rating} and \eqref{eq:osigmatau}.

\section{Implementation}
\label{sec:implementation}

In this section, we present a simple while efficient coordinate ascent method to solve the minimax program through its dual form and also a practical procedure for model selection.

\subsection{Coordinate Ascent}

The dual problem of regularized minimax conditional entropy for either multiclass or ordinal labels is nonconvex.  A stationary point can be obtained via coordinate ascent (Algorithm \ref{alg:em}), which is essentially Expectation-Maximization (EM) \citep{DemLaiRub77, NeaHin98}. We first  initialize the label estimate via aggregating votes in Equation \eqref{ag:mv}. Then, in each iteration step, given the current  estimate of the labels,  update the estimate of the confusion matrices of the workers and items by solving the optimization problem in \eqref{ag:mstep}; and,  given the current estimate of the confusion matrices of worker and item,  update the estimate of the labels through the closed-form formula in \eqref{ag:estep}, which  is identical  to applying the Bayes' rule with a uniform prior. The optimization problem in \eqref{ag:mstep} is strongly convex and smooth. Many  algorithms can be applied here \citep{YuNest2004}. In our experiments, we simply use gradient ascent. Denote by $F$ the objective function in \eqref{ag:mstep}.   For multiclass labels, the gradients are computed as
\begin{align}
\frac{\partial F}{ \partial \sigma_i (c, k)} &= \sum_{j} Q(Y_j = c)\left[\bbI(x_{ij} = k) - P(X_{ij} = k|Y_j = c)\right] - \alpha \sigma_i (c, k), \nonumber \\
\frac{\partial F}{ \partial \tau_j (c, k)} &= \sum_{i} Q(Y_j = c)\left[\bbI(x_{ij} = k) - P(X_{ij} = k|Y_j = c)\right] - \beta \tau_j (c, k).  \nonumber
\end{align}
For ordinal labels, the gradients are computed as
\begin{align}
\frac{\partial F}{ \partial \sigma_{is}^{\Delta, \nabla}} &= \sum_{c, k} \bbI(c \Delta s, k \nabla s)\sum_{j} Q(Y_j = c)\left[\bbI(x_{ij} = k) - P(X_{ij} = k|Y_j = c)\right] - \alpha \sigma_{is}^{\Delta, \nabla}, \nonumber \\
\frac{\partial F}{ \partial \tau_{js}^{\Delta, \nabla}} &= \sum_{c, k} \bbI(c \Delta s, k \nabla s)\sum_{i} Q(Y_j = c)\left[\bbI(x_{ij} = k) - P(X_{ij} = k|Y_j = c)\right] - \beta \tau_{js}^{\Delta, \nabla}.  \nonumber
\end{align}
It is worth pointing out that it is unnecessary to obtain the exact optimum at this intermediate step. We have observed that in practice, several gradient ascent steps here  suffice  for  reaching a final good solution.

\begin{algorithm}[t]
\caption{~Regularized Minimax Conditional Entropy for Crowdsourcing}
\label{alg:em}
\begin{algorithmic}
\STATE {\bfseries input:} $\{x_{ij}\}, \alpha, \beta $ \\[1.5ex]
\STATE {\bfseries initialize:}
\vspace{-1.0ex}
\begin{equation}
\label{ag:mv}
Q(Y_j = c) ~\propto~ \sum_{i} \bbI(x_{ij} = c)
\end{equation} 
\vspace{-4.0ex}
\STATE {\bfseries repeat:} 
\begin{subequations}
\begin{align}
& \{\sigma, \tau\} ~=~ \arg \max_{\sigma, \tau} \  \sum_{i, j, c} Q(Y_j = c)\log P(X_{ij} = x_{ij}|Y_j = c) - \alpha \Omega^*(\sigma) - \beta \Psi^*(\tau) \label{ag:mstep}\\[1.0ex]
& Q(Y_j = c)~\propto~\prod_i P(X_{ij} = x_{ij}|Y_j = c) \label{ag:estep}
\end{align}
\end{subequations}
\vspace{-2.5ex}
\STATE {\bfseries output:} $Q$
\end{algorithmic}
\end{algorithm}

\subsection{Model Selection}
\label{sec:cv}
The regularization parameters $\alpha$ and $\beta$ can be chosen as follows. If the true labels of a subset of items are known---such subsets are usually referred to as validation sets---we may choose the regularization parameters such that those known true labels can be best predicted.  Otherwise, we suggest to choose the regularization parameters via $k$-fold likelihood-based cross-validation. Specifically, we first randomly partition the crowd labels  into $k$ equal-size subsets,  and define a finite set of possible choices for the regularization parameters. Then, for each possible choice of the regularization parameters,
\begin{enumerate}
\item Leave out one subset and use the remaining $k-1$ subsets to estimate the confusion matrices  of the workers and items;
\item Plug the estimate into the probabilistic labeling model to compute the likelihood of the left-out subset;
\item Repeat the above two steps till each subset is left out once and only once;
\item Average the likelihoods that we have computed.
\end{enumerate}
After going through all the possible choices for the regularization parameters, we choose the one which results in the largest average
likelihood to run our algorithm over the full dataset.  The cross-validation parameter $k$  is typically set to 5 or 10.

To simplify the model selection process,  we suggest to choose
\begin{equation}
\label{eq:gamma}
\begin{aligned}
&\alpha = \gamma \times (\text{number of classes})^2, \\
&\beta = \frac{\text{number of labels per worker}}{\text{number of labels per item}} \times \alpha.
\end{aligned}
\end{equation}
In our experiments, we select  $\gamma$  from $\{2^{-2}, 2^{-1}, 2^0, 2^1, 2^2\}.$  In our limited empirical studies, larger candidate sets for $\gamma$  did not give more gains. Two empirical observations motivate us to consider using the square of the number of classes in Equation \eqref{eq:gamma}. First, the square of the number of classes has the same magnitude as the number of parameters in a confusion matrix.  Second, the label noise dramatically increases when the number of classes increases, requiring a super linearly scaled regularization.

\section{Related Work}
\label{sec:related}
In this section, we review some existing work that are closely related to our work.

 \textbf{Dawid-Skene Model}. Let $K$ denote the number of classes. \citet{DawSke79} propose a generative model in which the ability of worker $i$ is characterized by a $K \times K$ probabilistic confusion matrix $[p_i(c, k)]$ in which the diagonal element $p_i(c, c)$ represents the probability that  worker $i$ correctly labels an arbitrary item in class $c$, and the off-diagonal element $p_i(c, k)$ represents the probability that  worker $i$ mislabels an arbitrary item in class $c$ as class $k.$  Our probabilistic labeling model in Equation \eqref{eq:model} is reduced to the Dawid-Skene model when the item difficult terms $\tau_j(c, k)$ in our model disappear  since we can then reparameterize
 \[p_i(c, k) = \frac{\exp[\sigma_i(c, k)]}{\sum_{k'} \exp[\sigma_i(c, k')]}. \]
 In this sense, our model generalizes the Dawid-Skene model to incorporate item difficulty. To jointly estimate the workers' abilities and the true labels in the Dawid-Skene model,  in general, the marginal likelihood is maximized using the EM algorithm.

 For binary labeling task,  the probabilistic confusion matrix in the Dawid-Skene model can be written as
 \[
 \begin{pmatrix}
 p_i   & 1 - p_i\\
 1 - q_i  & q_i
 \end{pmatrix},
 \]
 where $p_i$ is the accuracy of worker $i$ in the first class, and $q_i$ the accuracy in the second class.  Usually,  this special case of the Dawid-Skene model is also referred to as the two-coin model \citep{RayYuZha10, LiuPenIhl12, CheLinZho13}. One may simplify the two-coin model by assuming $p_i = q_i$ \citep{GhoKalMca11,KarOhSha11,DalDasKum2013}.  This simplification is accordingly referred to as the one-coin model.

 \citet{KarOhSha11} propose an inference algorithm under the one-coin model,  and show that their algorithm achieves the minimax rate when the accuracy of every worker is bounded away from $0$ and $1, $ that is, with some fixed number $\epsilon > 0,  $ $\epsilon < p_i < 1 - \epsilon.$   \citet{LiuPenIhl12} show that the algorithm proposed by \citet{KarOhSha11} is essentially  a belief propagation update with the Haldane prior which assumes that each worker is either a hammer ($p_i = 1$) or adversary ($p_i = 0$) with equal probability.

 \citet{GaoZho14} show that  under the one-coin model, the global optimum of maximum likelihood achieves the minimax  rate.  A projected EM algorithm is suggested and  shown to achieve nearly the same rate as that of global optimum. \citet{ZhaCheZhoJor14} show that the EM algorithm for the general Dawid-Skene model can achieve the minimax rate up to a logarithmic factor when it is initialized by spectral methods \citep{AnaGeHsu12} and  the accuracy of every worker is  bounded away from $0$ and $1.$

 \citet{RayYuZha10} extend the Dawid-Skene model by imposing a beta prior over the worker confusion matrices. Moreover, they jointly learn the classifier and the true labels by assuming that the true labels are generated by a logistic model. \citet{LiuPenIhl12} develop full Bayesian inference via variational methods including belief propagation
and mean field.

\textbf{Rasch model \citep{Ras61,Ras68}.} In educational tests, the Rasch model illustrates the response of each examinee of a given ability to each item in a test. In the model, the probability of a correct response is modeled as a logistic function of the difference between the person and item parameter which are locations on a continuous latent trait. Person parameters represent the ability of examinees  while item parameters represent the difficulty of items.

Let $X_{ij} \in \{0, 1\}$ be a dichotomous random variable where $X_{ij} = 1$ denotes a correct response and $X_{ij} = 0$ an incorrect response to a given assessment item. Mathematically, the Rasch model is given by
\[P(X_{ij} = 1) = \frac{\exp(\beta_i - \delta_j)}{1 + \exp(\beta_i - \delta_j)}, \]
where $\beta_i$ is the ability of examinee  $i$ and $\delta_i$ the difficulty of item $j.$
The larger an examinee's ability relative to the difficulty of an item, the larger the probability of a correct response on that item. When the examinee's ability on the latent trait is equal to the difficulty of the item, the probability of a correct response is $1/2.$

The Rasch model is a special item response theory (IRT) model \citep{LorNov68}. However, unlike other IRT models,  the Rasch model satisfies the objective measurement principle pioneered by Rasch. Our work generalizes both the Rasch model and the objective measurement principle to multiclass labeling tasks. In addition, unlike the Rasch model, in our scenario, the true answers are unknown and have to be estimated.

\textbf{Polytomous Rasch model.}  The Rasch model has been adapted to the applications in which responses to items are scored with successive integers such as rating scales. Let $X_{ij} = \{0, 1, \cdots, m\}.$  \citet{AndRat1978} suggests
\[P(X_{ij} = k) = \frac{\exp \sum_{s=0}^k[\beta_i - (\delta_j - \tau_s)]}{\sum_{k'=0}^m \exp \sum_{s=0}^{k'}[\beta_i - (\delta_j - \tau_s)]}, \]
where $\beta_i$ is the location of person $i$ on a latent continuum, $\delta_j$  the difficulty of item $j$ on the same continuum, and $\tau_s$ the $s$-th threshold location of the rating scale which is in common to all the items. This model is usually referred to as the Rasch rating scale model. Later, the Rasch partial credit model  developed by \citet{MasARas1982} generalizes  the Rasch rating scale model into
\[P(X_{ij} = k) = \frac{\exp \sum_{s=0}^k(\beta_i - \tau_{js})}{\sum_{k'=0}^m \exp \sum_{s=0}^{k'}(\beta_i - \tau_{js})}, \]
where $\tau_{js}$ is the $s$-th threshold location of item $i$ on a latent continuum. When $\tau_{js}$ can be decomposed as $\tau_{js} = \delta_j - \tau_s, $ these two models coincide. \citet{uebersax1993latent} and \citet {MineiroOrdered11} apply the polytomous form of the Rasch model with minor changes to aggregate ordinal labels from a crowd.

\textbf{Probabilistic matrix factorization.} Let $X_{ij}$ be the label given by worker $i$ to item $j. $  Let $Y_j$ be the true label of item $j. $ \citet{WRWBM09} model the labeling process by revising the Rasch model into
\[P(X_{ij} = Y_j ) = \frac{1}{1 + \exp\left(-\dfrac{\beta_i}{\delta_j}\right)}, \]
and refer to their model as GLAD (Generative Model of Labels, Abilities, and Difficulties). It is easy to see that GLAD violates the principle of invariant comparison.  By using the per-worker confusion matrix in the Dawid-Skene model, \citet{MineiroConfusion11} generalizes GLAD to multiclass labeling as
\[P(X_{ij} = k|Y_j = c) \propto \exp\left[ \dfrac{\beta_{i}(c, k)}{\delta_j}\right]. \]
\citep{WBBP10}  parameterize workers and items with vectors and suggest
\[P(X_{ij} = Y_i) = \Phi({w}_i^\top {z}_j - b_j), \]
where $\Phi(\cdot)$ is the cumulative standardized normal distribution, ${w}_i\in \bbR^d$ the unobserved worker parameter,  and ${z}_j \in \bbR^d, b_j \in \bbR$  the unobserved item parameter. GLAD can be roughly thought of  as a special case of this model with  the dimension $d = 1 $ and $b_j = 0.$

\textbf{Other related work.} For other probabilistic modelling of crowdsourcing, we refer the readers to \citep{BacMinGui12,TiaZhu12,DaiLin13,VenGui14}. For online decision making in crowdsourcing, we refer the readers to \citep{ShePro08,abraham2013adaptive,CheLinZho13,SinKra13,HoSliWor14,AnaGoeNik14}. Regularized maximum entropy is studied in \citep{CheRos00,LebLaf01,KazTsu03,AltSmo06,DudPhiSch07}. \citet{zhwumu97} propose a minimax entropy method for  feature binding  and selection, and apply it to texture modeling and obtain a new class of Markov random field models. \citet{shaZhoDou14} propose a multiplicative payment mechanism to incentivize crowdsourcing workers to answer a question when they are sure and skip when they are not sure. They obtain extremely high quality crowdsourced data by using their  mechanism.

\section{Experiments}

\label{sec:exp}

In this section, we report empirical results of our method and some existing methods discussed in Section \ref{sec:related}.  Two error metrics are considered. One is the classification error rate for binary or multiclass data, and the other is  the mean square error for ordinal data.

\subsection{Datasets}
All datasets that we use are from real crowdsourcing tasks and publicly available.\footnote{Some of the datasets can be found at {http://research.microsoft.com/en-us/projects/crowd/}} The details are as follows:
\begin{itemize}
\item \textbf{Bluebirds \citep{WBBP10}}. This dataset contains a set of 108 images which are labeled as  indigo bunting or blue grosbeak by 39 crowdsourcing workers. Every worker labeled every image. The average error rate of the workers is $36.44\%,  $ compared to the error rate of random guessing at $50\%.$
\item \textbf{Price \citep{LiuSteIhl13}}.  This dataset consists of 80 household items collected from stores such as Amazon and Costco. The prices of the products are estimated by 155 undergraduate students from UC Irvine.
Seven price bins are created in this data collection:
          \$0$-$\$50,
          \$51$-$\$100,
          \$101$-$\$250,
         \$251$-$\$500,
         \$501$-$\$1000,
   \$1001$-$\$2000,
 and   \$2001$-$\$5000.
For each product, a student has to to decide which bin its price falls in. The average error rate of the students is $69.47\%$, compared to the error rate of random guessing at $85.71\%$. It may not be surprising that this dataset is systematically biased: all the students tend to underestimate the prices of the products.
\item \textbf{RTE \citep{SnoCon08}}. For each crowdsourced question,  the worker is presented with two sentences and asked to check if the second hypothesis sentence can be inferred from the first. This dataset contains 800 sentence pairs and 164 workers. Each sentence pair has 10 annotations. The average error rate of the workers is $15.87\%$, compared to the error rate of random guessing at $50\%$.
\item \textbf{Temp \citep{SnoCon08}}. For each crowdsourced question,  the worker is presented with a pair of verb events and asked to check if the event
described by the first verb occurs before or after the second.  This dataset contains 462 event pairs and 76 workers. Each event pair has 10 annotations. The average error rate of the workers is $16.30\%$, compared to the error rate of random guessing at $50\%$.
\item \textbf{Age \citep{HanOttLiuJai14}}. Amazon mechanical turkers are asked to estimate the age of a person in a face image. This dataset contains 1002 images and 165 workers. Each image has 10 age estimates.  Those estimates are integers not more than 100.  We put them into 7 bins: [1, 9], [10, 19], [20, 29], [30, 39], [40, 49], [50, 59], [60, 100].   With respect to this partition, the average error rate of the workers is $44.64\%$, compared to the error rate of random guessing at $85.71\%$.
\item \textbf{Web search \citep{zhoplaby12}}. This dataset contains 2665
query-URL pairs and 177 workers. Give a query-URL pair, a worker is required
to provide a rating to measure how the URL is relevant to the query. The
rating scale is 5-level: perfect, excellent, good, fair, or bad. On
average, each pair was labeled by around 6 different workers, and each worker
labeled around 90 pairs. More
than 10 workers labeled only one query-URL pair. The ground truth labels used for evaluation are obtained via a consensus among a group of 9 search experts.  The average error rate of the workers is
$62.95\%, $ compared to the error rate of random guessing at $80\%.$
\item \textbf{Web spam}. This dataset is provided by Microsoft web spam team.  It contains 149 web pages and 18 workers. The workers are required to identify which web pages are spam.  In average, each web page is labeled by  around 13 workers. The ground truth labels used for evaluation are provided by web spam experts. The average error rate of the workers is $16.30\%$, compared to the error rate of random guessing at $50\%$.
\end{itemize}
Table \ref{table:summary} shows a summary of these datasets.

\begin{table}[t]
\centering
\begin{tabular}{c|c|c|c|c}\hline
 & \# classes & \# items & \# workers & \# worker labels\\ \hline
 Bluebirds & $2$  & $108$ & $39$ & $4212$   \\ \hline
 Price & $7$ & $80$ & $155$ & $12400$ \\ \hline
 RTE & $2$  & $800$  & $164$ & $8000$   \\ \hline
 Temp & $2$  & $462$  & $76$ & $4620$   \\ \hline
 Age & $7$  & $1002$  & $165$ & $10020$   \\ \hline
 Web search & $5$  & $2665$ & $177$ & $15567$   \\ \hline
 Web spam & $2$  & $149$ & $18$  & $1901$   \\ \hline
\end{tabular}
\caption{Summary of the real crowdsourcing datasets used in our experiments. }
\label{table:summary}
\end{table}

\begin{table}[t]
\centering
\begin{tabular}{c|c|c|c|c|c}\hline
      & MV & DS-EM & DS-MF & GLAD &  MMCE(M)\\ \hline
 Bluebirds & $24.07$  & $10.19$ &$10.19$  & $12.04$ & $\mathbf{8.33}$   \\ \hline
 Price & $67.50$  & $\mathbf{65.00}$ &$67.50$  & $68.75$ & $67.50$   \\ \hline
 RTE & $10.31$  & $7.25$ &$\mathbf{6.63}$  & $7.00$ & $7.50$   \\ \hline
 Temp & $6.39$  & $5.84$ &$5.84$  & $\mathbf{5.63}$ & $\mathbf{5.63}$   \\ \hline
 Age & $34.88$  & $39.62$ &$36.33$  & $35.73$ & $\mathbf{31.14}$   \\ \hline
 Web search & $26.93$  & $16.92$ &$18.24$  & $19.30$ & $\mathbf{11.12}$   \\ \hline
 Web spam & $19.80$  & $13.42$ &$\mathbf{12.75}$  & $18.12$ & $\mathbf{12.75}$   \\ \hline
\end{tabular}
\caption{Error rates $(\text{in} \ \%)$  of various methods on real datasets. }
\label{table:errorrates}
\end{table}

\begin{table}[t]
\centering
\begin{tabular}{c|c|c|c|c|c|c}\hline
      & MV & DS-EM & DS-MF & LTA &  MMCE(M) & MMCE(O)\\ \hline
 Price & $1.605$  & $1.517$ &$1.487$  & $1.504$ & $1.643$ & $\mathbf{1.466}$    \\ \hline
 Age & $0.730$  & $0.852$ &$0.739$  & $0.696$ & $\mathbf{0.605}$ & $0.794$  \\ \hline
 Web search & $0.930$  & $0.539$ &$0.559$  & $0.481$ & $0.419$ & $\mathbf{0.384}$   \\ \hline
\end{tabular}
\caption{Mean square errors of various methods on ordinal datasets.    }
\label{table:meansqure}
\end{table}

\begin{table}[t]
\centering
\begin{tabular}{c|c|c|c|c|c|c}\hline
Probability Bin  & $(0, 0.5)$ & $(0.5, 0.6)$ & $(0.6, 0.7)$ & $(0.7, 0.8)$ & $(0.8, 0.9)$ & $(0.9, 1)$  \\ \hline
\# items  &$173$    &$291$    &$292$    &$313$    &$406$    &$1178$      \\ \hline
Error rate  &$0.416$  &$0.381$  &$0.199$  &$0.080$  &$0.020$  &$0.001$     \\ \hline
Mean square error  &$0.832$  &$0.423$  &$0.250$  &$0.118$  &$0.035$  &$0.001$     \\ \hline
\end{tabular}
\caption{Positive correlation between probabilistic labels and errors. The results are from the regularized ordinal minimax conditional entropy method on the web dataset.}
\label{table:probcor}
\end{table}

\subsection{Methods}
We evaluate the following methods in our experiments:
\begin{itemize}
\item \textbf{Majority voting (MV)}. It is perhaps the simplest baseline.
\item \textbf{Dawid-Skene model $+$ EM (DS-EM)}. Under the generative model by \citep{DawSke79}, this method jointly estimates workers' parameters and true labels by maximizing the likelihood of observed labels with the EM algorithm.
\item \textbf{Dawid-Skene model $+$ mean field (DS-MF)}. This method performs variational Bayesian inference using the mean field (MF) algorithm \citep{LiuPenIhl12}. It assumes a Dirichlet prior parameterized by a vector $\alpha_k$ on the $k$-th row of the worker confusion matrix in the Dawid-Skene model with $\alpha_{k,k} = c_1$, and $\alpha_{k,l}= c_2$ for all $l\neq k$. The hyperparameters $\{c_1, c_2\}$ are selected by maximizing the marginal likelihood calculated by MF, and searched in a $10\times 10$ grid defined by $c_1 = c_2 \times \{10^{0}, 10^{0.1}, 10^{0.2}, \cdots, 10^1\}$ and $c_2 = \{10^{-1}, 10^{-0.8}, \cdots, 10^0, \cdots, 10^{0.8}, 10^1\}$.
\item \textbf{GLAD}. We use the multiclass version of GLAD proposed by \citep{MineiroConfusion11} and also his open source implementation.
\item \textbf{Latent trait analysis (LTA)}. It is a variant of the polytomous Rasch model proposed by  \citep{MineiroOrdered11} with an  open source implementation.
\item \textbf{Regularized minimax conditional entropy for multiclass labels (MMCE(M))}. It is implemented with the Euclidian norm based regularization.
\item \textbf{Regularized minimax conditional entropy for ordinal labels (MMCE(O))}. It is implemented with the Euclidian norm based regularization.
\end{itemize}
The regularization parameters in MMCE are chose through the cross-validation procedure described in Section \ref{sec:cv}.\footnote{Our code are available at {http://research.microsoft.com/en-us/projects/crowd/}}

\subsection{Results}
Table \ref{table:errorrates} shows the error rates of various methods on real crowdsourcing datasets. Our multiclass minimax conditional entropy method outperforms compared methods on most datasets.
Table \ref{table:meansqure} shows the mean square errors of various methods on three ordinal datasets. Our ordinal minimax conditional entropy method performs best on the  price and web search datasets but performs poorly on the age dataset. Table \ref{table:probcor} shows the correlation between probabilistic labels and errors for our ordinal minimax conditional entropy method on the web dataset. From the results, the labels estimated with larger probabilities are more likely to be correct. We observed similar behavior  for our multiclass minimax conditional entropy method. We also evaluated our method with the regularization in Equation \eqref{eq:comg} and observed that this variant  somewhat hurts performance on most datasets.

\section{Conclusion}

\label{sec:conclusion}

We have developed a minimax conditional entropy principle for aggregating noisy labels from crowdsourcing workers. Our formulation involves two probabilistic distributions. One is the distribution of the true labels of the items, and the other is the distribution under which the workers generate their labels for the items. Both the distributions are unknown. We jointly infer them by first maximizing the entropy of the observed labels of the workers  conditioned on the true labels of the items over the distribution of generating workers' labels,  and then minimizing the maximum entropy over the distribution of the true labels of the items. Empirical results on real crowdsourcing datasets validate our approach.

 We have considered aggregating multiclass and ordinal labels via minimax conditional entropy. The framework is  general and should be extensible  to many other labeling tasks in which the labels are structured in different ways, such as  protein folding \citep{khatib2011algorithm}, machine translation \citep{zaidan2011crowdsourcing}, hierarchical classification \citep{KolSah}, and speech captioning \citep{MurMilLas2013}. To achieve the extension,   the constraints for workers and items  need to be customized specific to each domain,  and this probably results in differently structured confusion matrices.

\subsection*{Acknowledgements}
{We would like to thank  Sumit Basu and Yi Mao for their early contribution to this work, Daniel Hsu, Xi Chen, Chris Burges for helpful discussions, and Gabriella
Kazai for providing  the web search dataset. }


\newpage

\appendix

\section{Dual Form of  Minimax Conditional Entropy}
\label{app:dual1}
To to derive the dual of minimax conditional entropy, we substitute the probabilistic model in Equation \eqref{eq:model} into the Lagrangian \eqref{eq:lag1} and obtain
\begin{align*}
L = & -\sum_{i, j, c}  Q(Y_j = c)\sum_k P(X_{ij} = k| Y_j = c)\log \bigg\{\frac{1}{Z_{ij}}\exp[\sigma_i(c, k) + \tau_{j}(c, k)]\bigg\} \\
  & + \sum_{i, c, k} \sigma_i(c, k) \sum_j Q(Y_j = c)\bigg[ P(X_{ij} = k |Y_j = c) -  \bbI(x_{ij} = k) \bigg]\\
  & + \sum_{j, c, k} \tau_j(c, k)   \sum_i Q(Y_j = c) \bigg[P(X_{ij} = k |Y_j = c)-  \bbI(x_{ij} = k) \bigg]\\
   & + \sum_{i,j, c} \lambda_{ijc}\bigg[\sum_{k} P(X_{ij} = k|Y_j = c) -1 \bigg]\\
   = & -\sum_{i, j, c}  Q(Y_j = c)\sum_k P(X_{ij} = k| Y_j = c)[\sigma_i(c, k) + \tau_{j}(c, k)] + \sum_{i, j} \log  Z_{ij}\\
   & + \sum_{i, c, k} \sigma_i(c, k) \sum_j Q(Y_j = c)\bigg[ P(X_{ij} = k |Y_j = c) -  \bbI(x_{ij} = k) \bigg]\\
  & + \sum_{j, c, k} \tau_j(c, k)   \sum_i Q(Y_j = c) \bigg[P(X_{ij} = k |Y_j = c)-  \bbI(x_{ij} = k) \bigg]\\
  =  & -\sum_{i, j, c}  Q(Y_j = c)\bigg(\sum_k \bbI(x_{ij} = k)[\sigma_i(c, k) + \tau_{j}(c, k)] -  \log  Z_{ij}\bigg)\\
  =  & -\sum_{i, j, c}  Q(Y_j = c)\log P(X_{ij} = x_{ij}|Y_j = c).
\end{align*}

\section {Proof of Theorem \ref{eq:minkl}}
\label{sec:proofkl}
Let us first check $\sum_{X, Y}Q(X,Y) \log {Q(X,Y)}. $ By definition,
\[\sum_X Q(X) \log Q(X) = 0 , \quad  Q(X|Y) = Q(X). \]
Hence, we have
\begin{align}
 \sum_{X, Y}Q(X,Y) \log {Q(X,Y)} & = \sum_{X, Y}[Q(X|Y)Q(Y)] \log [Q(X|Y)Q(Y)] \nonumber\\
 & = \sum_{X, Y}[Q(X)Q(Y)] \log [Q(X)Q(Y)] \nonumber\\
 & = \sum_X Q(X) \log Q(X) + \sum_Y Q(Y)\log Q(Y) \nonumber \\
 & = \sum_Y Q(Y)\log Q(Y). \nonumber
\end{align}
Next we check  $\sum_{X, Y}Q(X,Y) \log {P(X,Y)}. $ Write
 \begin{equation*}
 \begin{aligned}
\sum_{X, Y} Q(X, Y) \log P(X, Y) = \sum_{X, Y} Q(X, Y) \log P(X|Y)+ \sum_{X, Y} Q(X, Y) P(Y).
 \end{aligned}
\end{equation*}
Since $P$ is a uniform distribution over $Y,$ $P(Y)$ is a constant. Thus,
\begin{align}
\sum_{X, Y} Q(X, Y) \log P(Y)  & = \log P(Y) \sum_{X, Y} Q(X, Y) =  \log P(Y), \nonumber
\end{align}
which is still a constant.  By Equation \eqref{eq:model}, we have 
\begin{equation*}
\begin{aligned}
& \sum_{X, Y} Q(X, Y) \log P(X|Y)   =   \sum_{i,j,c, k} Q(X_{ij} = k, Y_j = c) \log P(X_{ij} = k| Y_j = c) \\
= & \sum_{i,j,c, k} Q(X_{ij} = k, Y_j = c) \log \bigg\{\frac{1}{Z_{ij}} \exp[\sigma_i(c, k) + \tau_{j}(c, k)]\bigg\} \\
 = & \sum_{i,j,c, k} Q(X_{ij} = k, Y_j = c)  \left[\sigma_i (c, k) +  \tau_j (c, k)- \log Z_{ij}\right].
\end{aligned}
\end{equation*}
By Equation \eqref{eq:mct1}, we have
\begin{equation*}
\begin{aligned}
 \sum_{i,j,c, k} Q(X_{ij} = k, Y_j = c) \sigma_i (c, k)  = & \sum_{i,c,k} \sigma_i (c, k) \sum_{j}\bbI(X_{ij} = k) Q( Y_j = c)  \\
= & \sum_{i,c,k} \sigma_i (c, k) \sum_j  P(X_{ij} = k |Y_j = c) Q(Y_j = c) . \\
\end{aligned}
\end{equation*}
Similarly, by Equation \eqref{eq:mct2},
\[\sum_{i,j,c, k} Q(X_{ij} = k, Y_j = c) \tau_{j}(c, k) = \sum_{j,c,k} \tau_j (c, k) \sum_i   P(X_{ij} = k|Y_j = c)Q(Y_j = c).  \]
In addition, since $Z_{ij}$ does not depend on  $k,$
\begin{equation*}
\begin{aligned}
 \sum_{i,j,c, k}Q(X_{ij}  = k, Y_j = c) \log Z_{ij}  = & \sum_{i,j} \sum_c \log Z_{ij} \sum_{k}Q(X_{ij} = k , Y_j = c) \\
 = & \sum_{i,j}\sum_{c} Q(Y_j=c)\log Z_{ij} \\
 = & \sum_{i,j}\sum_{ c} Q(Y_j=c)\sum_{k} {P(X_{ij} = k|Y_j = c)}\log Z_{ij} .
\end{aligned}
\end{equation*}
Putting all the pieces together, we have
\begin{equation*}
\begin{aligned}
D_{\KL}(Q\parallel P) = & -\sum_{j, c} Q(Y_j = c) \sum_{i,k} P(X_{ij} = k|Y_j = c)[\sigma_i (k,c)  + \tau_j (c, k) - \log Z_{ij}]\\
  &  + \sum_{Y} Q(Y)\log Q(Y)-  \log P(Y)\\
 = & - \sum_{j, c} Q(Y_j = c) \sum_{i, k} P(X_{ij} = k|Y_j = c)\log P(X_{ij} = k|Y_j = c) \\
 &  + \sum_{Y} Q(Y)\log Q(Y) -  \log P(Y). \\
\end{aligned}
\end{equation*}
Note that, when the true labels are deterministic,
\[\sum_{Y} Q(Y)\log Q(Y) = 0. \]
So,
\[D_{\KL}(Q\parallel P) = -\sum_{j, c} Q(Y_j = c) \sum_{i, k} P(X_{ij} = k|Y_j = c)\log P(X_{ij} = k|Y_j = c)-  \log P(Y) .  \]
This concludes the proof.

\section{Dual Form of  Regularized Minimax Conditional Entropy}
\label{app:dual2}
We derive the dual problem of regularized maximum conditional entropy with the sum-to-zero constraints in Equation \eqref{eq:sumtozero}. The dual derivation without the additional constraints can be obtained in a similar procedure. Let us write the Lagrangian as
\begin{equation}
\label{eq:plag}
L = H(X|Y) - H(Y)- \frac{1}{\alpha}\Omega(\xi)  - \frac{1}{\beta}\Psi(\zeta) + L_\sigma + L_\tau + L_{\lambda} + L _{\mu},
\end{equation}
in which
\begin{align*}
& L_{\sigma}　= \sum_{i, c, k} \sigma_i(c, k)\bigg[\xi_{i}(c, k) - \sum_{j}\left(\phi_{ij}(c, k) - \widehat{\phi}_{ij}(c, k)\right)\bigg],\\
& L_{\tau} = \sum_{j, c, k} \tau_j(c, k)\bigg[\zeta_{j}(c, k) - \sum_i \left(\phi_{ij}(c, k) - \widehat{\phi}_{ij}(c, k)\right)\bigg],\\
& L_{\lambda} = \sum_{i,j, c} \lambda_{ijc}\bigg[\sum_{k,c} P(X_{ij} = k|Y_j = c) -1 \bigg],  \\
& L _{\mu} = \sum_{i}\mu_i \sum_{c} \xi_i(c, c).
\end{align*}
By the KKT conditions, maximizing $L$ with respect to $P$ results in
\begin{align*}
\frac{\partial L }{\partial P(X_{ij} = k|Y_j = c)} = -\log P(X_{ij} = k|Y_j = c)-1 + \lambda_{ijc}  + \sigma_i(c, k)  + \tau_{j}(c, k) = 0.
\end{align*}
As showed in Section \ref{sec:cat}, this leads to the probabilistic model in Equation \eqref{eq:prob1}. Similarly, maximizing $L$ with respect to $\xi$ results in
\begin{align*}
& \frac{\partial L}{\partial \xi_i (c, k)} = \sigma_i (c, k) - \frac{1}{\alpha} \xi_i (c, k) = 0,  \ \forall c, k \neq c, \\
& \frac{\partial L}{\partial \xi_i (c, k)} = \sigma_i (c, c) - \frac{1}{\alpha} \xi_i (c, c) + \mu_i = 0, \ \forall c.
\end{align*}
So we have
\begin{align}
& \xi_i (c, k) = \alpha \sigma_i (c, k), \ \forall c, k \neq c, \label{eq:pxi1}\\
&  \xi_i (c, c) = \alpha [\sigma_i (c, c) + \mu_i], \ \forall c. \label{eq:pxi2}
\end{align}
Moreover, maximizing $L$ with respect to $\zeta$ results in
\begin{align*}
\frac{\partial L}{\partial \zeta_i (c, k)} = \tau_i (c, k) - \frac{1}{\beta} \zeta_i (c, k) = 0,  \ \forall c, k.
\end{align*}
Hence,
\begin{equation}
\zeta_i (c, k)  = \beta \tau_i (c, k), \ \forall c, k.  \label{eq:pzeta}
\end{equation}
Substituting \eqref{eq:model}, \eqref{eq:pxi1}, \eqref{eq:pxi2}, and \eqref{eq:pzeta} into  the Lagrangian \eqref{eq:plag}, we have
\begin{align*}
L = & -\sum_{i, j, c}  Q(Y_j = c) \log \bigg\{\frac{1}{Z_{ij}}\exp[\sigma_i(c, x_{ij}) + \tau_{j}(c, x_{ij})]\bigg\} - H(Y) \\
& - \frac{\alpha}{2} \sum_{i, c}\bigg\{\sum_{k \neq c} [\sigma_i(c, k)]^2 + [\sigma_i(c, c) + \mu_i]^2\bigg\}- \frac{\beta}{2} \sum_{i, c, k} [\tau_j(c, k)]^2 \\
& + \alpha \sum_{i, c} \bigg\{\sum_{k \neq c} [\sigma_i(c, k)]^2 +  \sigma_i(c, c) [\sigma_i(c, c)   + \mu_i]\bigg\} +  \beta \sum_{i, c, k} [\tau_j(c, k)]^2 \\
&  + \alpha \sum_{i, c}\mu_i(c, c) [\sigma_i(c, c)   + \mu_i]\\
= & -\sum_{i, j, c}  Q(Y_j = c)\log \bigg\{\frac{1}{Z_{ij}}\exp[\sigma_i(c, k) + \tau_{j}(c, k)]\bigg\} - H(Y) \\
& +  \frac{\alpha}{2} \sum_{i, c}\bigg\{\sum_{k \neq c} [\sigma_i(c, k)]^2 + [\sigma_i(c, c) + \mu_i]^2\bigg\} +  \frac{\beta}{2} \sum_{i, c, k} [\tau_j(c, k)]^2.
\end{align*}
By minimizing the Lagrangian over $\mu_i, $ we obtain
\[\mu_i = - \overline{\sigma_i(c, c)}. \]
So,  the dual problem can be expressed as
\begin{align*}
\min_{Q, \sigma, \tau} \quad & -\sum_{i, j, c}  Q(Y_j = c)\log \bigg\{\frac{1}{Z_{ij}}\exp[\sigma_i(c, x_{ij}) + \tau_{j}(c, x_{ij})]\bigg\} - H(Y) \\
& +  \frac{\alpha}{2} \sum_{i, c}\bigg\{\sum_{k \neq c} [\sigma_i(c, k)]^2 + \left[\sigma_i(c, c) - \overline{\sigma_i(c, c)}\right]^2\bigg\} +  \frac{\beta}{2} \sum_{i, c, k} [\tau_j(c, k)]^2.
\end{align*}
Let us replace $\sigma_i(c, k) $ with $\sigma_i(c, k) + \nu_{i}.  $  It is easy to verify  that this dual problem can be equivalently written as
\begin{align*}
\min_{Q, \sigma, \tau, \nu} \quad & -\sum_{i, j, c}  Q(Y_j = c)\log \bigg\{\frac{1}{Z_{ij}}\exp[\sigma_i(c, x_{ij}) + \tau_{j}(c, x_{ij})]\bigg\} - H(Y) \\
& +  \frac{\alpha}{2} \sum_{i, c}\bigg\{\sum_{k \neq c} [\sigma_i(c, k) + \nu_{i} ]^2 + \left[\sigma_i(c, c) - \overline{\sigma_i(c, c)}\right]^2\bigg\} +  \frac{\beta}{2} \sum_{i, c, k} [\tau_j(c, k)]^2.
\end{align*}
Minimizing the objective function over $\nu$ leads to
\begin{align*}
\min_{Q, \sigma, \tau} \quad & -\sum_{i, j, c}  Q(Y_j = c)\log \bigg\{\frac{1}{Z_{ij}}\exp[\sigma_i(c, x_{ij}) + \tau_{j}(c, x_{ij})]\bigg\} - H(Y) \\
& +  \frac{\alpha}{2} \sum_{i, c}\bigg\{\sum_{k \neq c} \left[\sigma_i(c, k) -  \overline{\sigma_i(c, k)} \right]^2 + \left[\sigma_i(c, c) - \overline{\sigma_i(c, c)}\right]^2\bigg\} +  \frac{\beta}{2} \sum_{i, c, k} [\tau_j(c, k)]^2.
\end{align*}

\section{Coordinate Algorithm}
\label{app:cm}
To solve the dual problem
\[\max_{\sigma, \tau, Q} \quad   \sum_{j, c} Q(Y_j = c)\sum_i \log P(X_{ij} = x_{ij}|Y_j = c) + H(Y) - \alpha \Omega^*(\sigma) - \beta \Psi^*(\tau) \]
subject to the probability constraints
\[ \sum_c Q(Y_j = c) = 1, \forall j, \ Q(Y_j = c) \ge 0,  \ \forall j, k, \]
we first split the variables into two groups and then alternatively update them. One group contains the parameters of workers and items in $P(X_{ij} = x_{ij}|Y_j = c), $ that is, $\{\sigma_{i}(c, k), \tau_j(c, k), \forall i, j, c, k\}$
and the other groups contains the unknown true labels $\{Q(Y_j = k), \forall j, k\}.$  When we update the variables in the first group, the variables in the second group take their current values. Then, the optimization problem becomes
\[\max_{\sigma, \tau} \quad   \sum_{j, c} Q(Y_j = c)\sum_i \log P(X_{ij} = x_{ij}|Y_j = c)  - \alpha \Omega^*(\sigma) - \beta \Psi^*(\tau). \]
Instead, when we update the variables in the second group, the variables in the first group take their current values. We thus have the optimization problem
\[\max_{Q} \quad   \sum_{j, c} Q(Y_j = c)\sum_i \log P(X_{ij} = x_{ij}|Y_j = c) + H(Y)\]
subject to the above probability constraints. This constrained optimization problem can be solved with the Lagrangian dual
\[L = \sum_{j, c} Q(Y_j = c)\sum_i \log P(X_{ij} = x_{ij}|Y_j = c) + H(Y) - \sum \lambda_{j} \bigg[\sum_c Q(Y_j = c) - 1\bigg],  \]
where $\lambda_{j}$'s are the Lagrangian multipliers.  By the KKT conditions,
\[\frac{\partial L}{\partial Q(Y_j = c)} = \sum_i \log P(X_{ij} = x_{ij}|Y_j = c) - \log Q(Y_j = c) + 1  - \lambda_{j} = 0.    \]
This implies
\[Q(Y_j = c) \propto \prod_{i} P(X_{ij} = x_{ij}|Y_j = c). \]



%
%
%

\vskip 0.2in
\bibliographystyle{plainnat}
\bibliography{crowd}

\end{document}